\documentclass[10pt,journal]{IEEEtran}

\usepackage{amsmath,amssymb,amsfonts}
\usepackage{graphicx}
\usepackage{subfigure}
\usepackage[hyphens]{url}
\usepackage{multirow}
\usepackage{float}
\usepackage{booktabs}
\usepackage{pifont}
\usepackage{todonotes}
\usepackage{adjustbox}
\usepackage{threeparttable}
\usepackage{algorithm}
\usepackage{algorithmic}
\usepackage{makecell}

\usepackage[hidelinks]{hyperref}
\hypersetup{
pdfauthor={J. Priesnitz, C. Rathgeb, N. Buchmann, C. Busch},
pdfsubject={Biometrics},
pdftitle={SynCoLFinGer: Synthetic Contactless Fingerprint Generator},
pdfkeywords={Mobile Biometrics, Fingerprint Recognition, Contactless Fingerprint, Synthetic Generation},
bookmarks=true,
bookmarksopen=true,
bookmarksnumbered=true,
pagebackref=false
}

\ifCLASSOPTIONcompsoc
  \usepackage[nocompress]{cite}
\else
  \usepackage{cite}
\fi

\def\eg{\textit{e.g}. } 
\def\ie{\textit{i.e}. } 
\def\cf{\textit{c.f}. }

 \def\etal{\textit{et al}. }

\begin{document}
\title{SynCoLFinGer:\\ Synthetic Contactless Fingerprint Generator}

\author{\IEEEauthorblockN{
        Jannis~Priesnitz$^*$,
        Christian~Rathgeb$^*$,
        Nicolas~Buchmann$^\dagger$,
        Christoph~Busch$^*$} \\ \vspace{0.1cm}
    \IEEEauthorblockA{$^*$Hochschule Darmstadt, Germany\\$^\dagger$Freie Universität Berlin, Germany
     \vspace{0.1cm}\\\small{\texttt{jannis.priesnitz@h-da.de}}
} \vspace{-0.5cm}
}

\IEEEtitleabstractindextext{
\begin{abstract}
We present the first method for synthetic generation of contactless fingerprint images, referred to as SynCoLFinGer. To this end, the constituent components of contactless fingerprint images regarding capturing, subject characteristics, and environmental influences are modeled and applied to a synthetically generated ridge pattern using the SFinGe algorithm. The proposed method is able to generate different synthetic samples corresponding to a single finger and it can be parameterized to generate contactless fingerprint images of various quality levels. The resemblance of the synthetically generated contactless fingerprints to real fingerprints is confirmed by evaluating biometric sample quality using an adapted NFIQ 2.0 algorithm and biometric utility using a state-of-the-art contactless fingerprint recognition system.
\end{abstract}

\begin{IEEEkeywords}
Mobile Biometrics, Fingerprint Recognition, Contactless Fingerprint, Synthetic Generation
\end{IEEEkeywords}}

\maketitle

\IEEEdisplaynontitleabstractindextext
\IEEEpeerreviewmaketitle

	\section{Introduction}
	Biometric system development and evaluation can benefit from the use of synthetic biometric data \cite{Buettner2009}. On the one hand, synthetic biometric data can be used for algorithm training where the acquisition of real samples is cost and time intensive. On the other hand, biometric systems can be evaluated with easily manageable synthetic data, which is generally not limited by data protection regulations. 

In the scientific literature, two main approaches for synthetic biometric data generation can be distinguished, \textit{modeling} and \textit{learning-based} approaches. Biometric signals or features may be modeled using hand-crafted methods specifically designed for biometric characteristics. Such models usually require knowledge about statistical properties of biometric data and have been proposed for various biometric characteristics. Such approaches have been carried out for several modalities including 3D contactless fingerprints \cite{labati2012virtual, long20153d}, contact-based fingerprints \cite{Cappelli2009} or finger vein \cite{Hillerstroem14}. Learning-based approaches which are commonly based on Generative Adversarial Network (GAN) \cite{Goodfellow14} architectures have been found to be suitable to generate realistic image data including biometric samples, \eg face \cite{Karras19} or iris images \cite{Yadav19}. Recent works also show competitive results in the area of contact-based fingerprint images \cite{engelsma2022printsgan, seidlitz2021generation}.


Focusing on fingerprint recognition \cite{Maltoni-HandbookOfFingerprintRecognition-2009}, some approaches to synthetically generate contact-based fingerprints have been proposed \cite{Cappelli2009, engelsma2022printsgan}. Pioneer work in this field has been done by Cappelli~\etal~\cite{Cappelli20} who proposed Synthetic Fingerprint Generator (SFinGe). Starting from the positions of cores and deltas, the SFinGe algorithm exploits a mathematical flow model to generate a consistent directional map. Subsequently, a density map is created on the basis of some heuristic criteria and the ridge-line pattern. Furthermore, the minutiae are created through a space-variant linear filtering; the output is a near-binary clear fingerprint image. Finally, specific noise is added to produce a realistic grayscale representation of the fingerprint. The latter step allows for the generation of multiple mated samples of a single finger. Up until now, continuous improvements have been applied to the SFinGe algorithm\footnote{Biometric System Laboratory -- University of Bologna: \url{http://biolab.csr.unibo.it/}}. Similarly, further approaches to model contact-based fingerprints have been proposed by different research laboratories, \eg in \cite{Zhao12}. 


\begin{figure}[t]	
	\centering
	\subfigure[real]{\includegraphics[height=4.8cm]{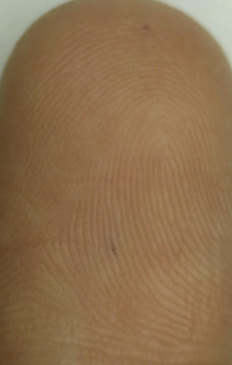}}\hfil
	\subfigure[SynCoLFinGer]{\includegraphics[height=4.8cm]{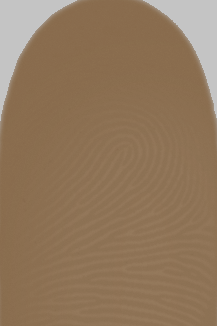}}
	\caption{Examples of a real and synthetic contactless fingerprint sample: (a) contactless sample from the ISPFDv2 database (cropped), (b) contactless synthetic sample generated by the proposed method.}
	\label{fig:fingerprint_examples}\vspace{-0.3cm}
\end{figure}

As mentioned before, GAN-based approaches have been employed for synthetically generating contact-based fingerprints, \eg in \cite{engelsma2022printsgan,seidlitz2021generation,riazi2020synfi,Fahim20}. 

However, several training data related challenges have to be solved to implement a GAN-based generation of contactless fingerprint images with reasonable realism and diversity. GANs require huge training databases. 
    To the best of the authors' knowledge a database of sufficient size is not publicly available for contactless fingerprint recognition. GANs learn the properties of the training database. Consequently, the generated images reflect the training data which may lead to a certain bias in case the training database is lacking diversity \cite{drozdowski2020demographic}. This also holds for non-biometric aspects such as environmental influences. 
    Moreover, GANs often fail to robustly generate  mated-samples. The reason for this is that the GANs have to learn to distinguish between persistent and non-persistent characteristics. Here, database size and heterogeneity is crucial to achieve realistic results. 
To the best of the authors' knowledge there is no contactless fingerprint database available which covers said aspects. Hence, a GAN-based synthetic contactless fingerprint generation is out of scope of this work.

Besides contact-based fingerprint recognition, the development of contactless systems represents a growing research area  \cite{labati2014touchless,priesnitz2021overview}. Contactless fingerprint capturing devices overcome some problems of contact-based ones like latent fingerprints of previous users (ghost fingerprints) or hygienic concerns. Since the contactless fingerprint capturing process is  different compared to the contact-based process, the resulting biometric samples are also different. Contactless fingerprints do not exhibit an elastic deformation caused by pressing the finger on a sensor plate. Moreover, many contactless fingerprint capturing devices are based on commodity equipment like smartphones, which produce RGB images. 
	
Despite the growing interest in contactless fingerprint recognition, so far no approaches to synthetic contactless fingerprint generation have been published. In this work, we propose the first (to the best of the authors' knowledge) method for generating synthetic contactless 2D fingerprint images named SynCoLFinGer. Our method generates fully synthetic images of a fingertip which have a similar appearance as pre-segmented fingerprint images captured by a contactless device, \textit{e.g.} smartphone. SynCoLFinGer includes the most common properties of contactless fingerprint imaginary such as the lack of elastic deformations of the fingerprint, a natural finger color and regions of low contrast.

Starting from a synthetically generated ridge pattern (using the SFinGe software \cite{Cappelli2009}), the constituent components of contactless fingerprint images with respect to capturing conditions, subject characteristics, as well as environmental influences are modeled to generate synthetic contactless fingerprint samples. By varying certain changeable parameters biometric variance and variation in sample quality are simulated which enables the  generation of various mated synthetic samples corresponding to a single finger. The generated contactless fingerprint samples exhibit good visual resemblance to real images, see Fig.~\ref{fig:fingerprint_examples} for an illustration. Furthermore, the realism of the synthetically generated contactless fingerprints is verified by estimating  sample quality employing an adapted NFIQ 2.0 algorithm and biometric utility using a state-of-the-art contactless fingerprint recognition system. To facilitate reproducible research and future experiments, the source code of SynCoLFinGer will be made publicly available\footnote{SynCoLFinGer source code available at:\newline https://gitlab.com/jannispriesnitz/syncolfinger
}.



	\begin{figure*}
		\centering
		\includegraphics[width=0.999\linewidth]{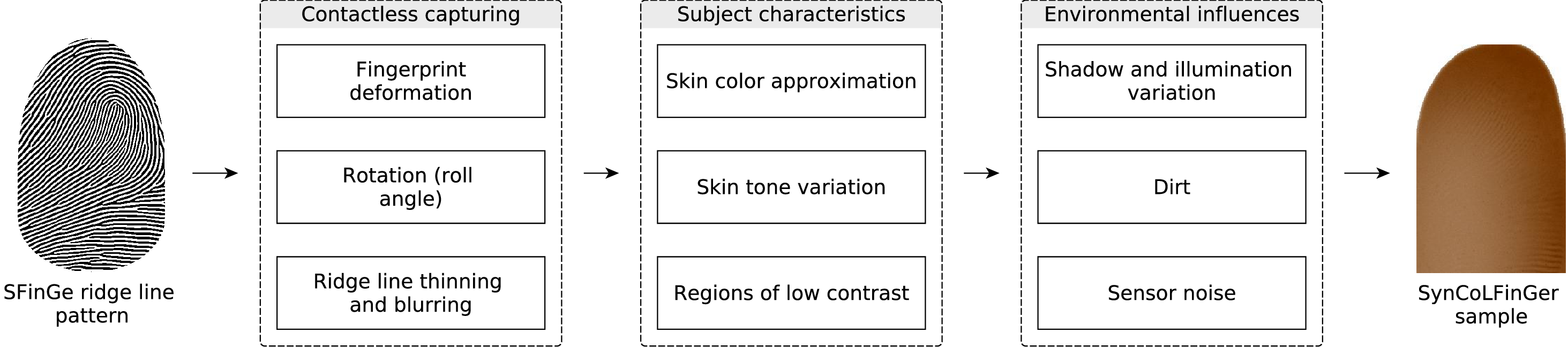}
		\caption{Overview of SynCoLFinGer. First step: simulation of a contactless capturing, second step: generation of subject-related characteristics, third step: simulation of environmental influences.}
		\label{fig:main_workflow}\vspace{-0.5cm}
	\end{figure*}
	
	\begin{figure}[t]	
		\centering
		\subfigure[SFinGe ridge pattern ]{\includegraphics[height=6.5cm]{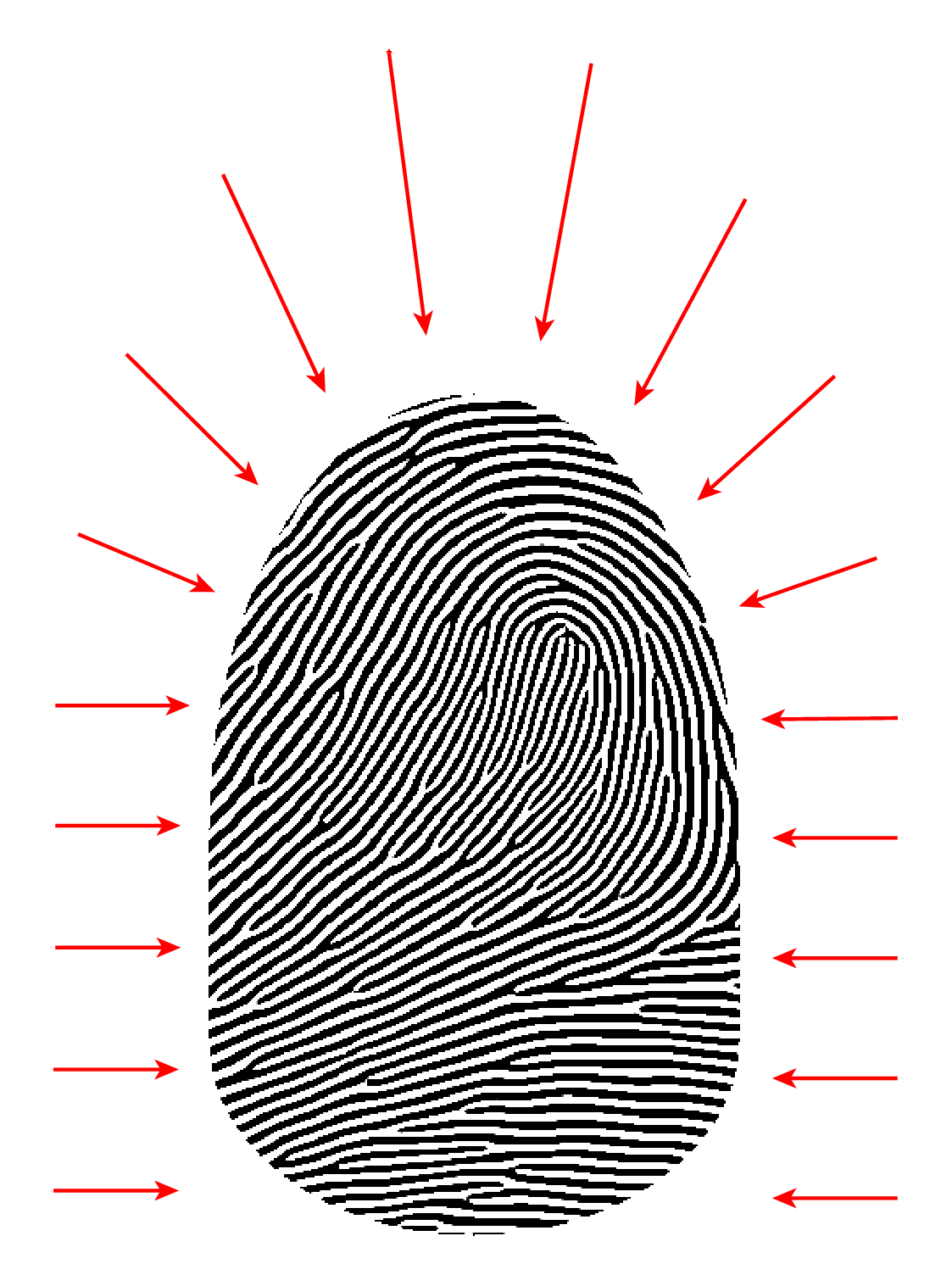}\hspace{0.8cm}}
		\subfigure[Contactless ridge pattern]{\includegraphics[height=4cm]{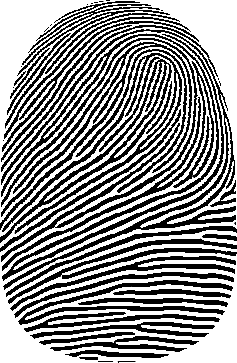}}
		\caption{Visualization of the fingerprint deformation with deformation vectors shown in red.}
		\label{fig:deformationfigure}
	\end{figure}

This work is organized as follows: the proposed synthetic contactless fingerprint generation method is described in detail in Sect.~\ref{sec:system}. Experimental evaluations are presented in Sect.~\ref{sec:experiments}. Finally, conclusions are drawn in Sect.~\ref{sec:conclusion}. 

	\section{Proposed Method}\label{sec:system}
	The proposed SynCoLFinGer method aims at generating a segmented fingertip. For this reason, the detection and segmentation of a fingertip and the rotation into an upright position is considered as solved processing steps. This is a reasonable assumption considering the detection and segmentation performance of state-of-the-art algorithms, \eg \cite{priesnitz2021deep}. Our method utilizes the well-known contact-based fingerprint generation algorithm SFinGe \cite{Cappelli2009}. For our method we use an intermediate result of the SFinGe algorithm which generates only the ridge line characteristic of a 500 dpi live scanned fingerprint but applies no further processing to the sample. 
	
	We apply three main steps to the SFinGe ridge pattern (see Fig. \ref{fig:main_workflow}): 
	\begin{enumerate}
		\item Application of a deformation function $d$ to a contact-based ridge line pattern $\mathbf{r}_{\mathit{cb}}$ to simulate a ridge line pattern $\mathbf{r}_{\mathit{cl}}$ of a contactless capturing scenario (see Sect.~\ref{sec:capturing}).
		\item Addition of subject-related characteristics $\mathbf{c}_s$ to the contactless ridge line pattern $\mathbf{r}_{\mathit{cl}}$ (see Sect.~\ref{sec:subject}). 
		\item Addition of a set of environmental characteristics $\{\mathbf{c}_{e_i}\}_{i=1}^n$ to generate a set of $n$ mated synthetic contactless fingerprints $\{\mathbf{f}_i\}_{i=1}^n$ (see Sect.~\ref{sec:environment}).
	\end{enumerate}
	
	The generation process of a SynCoLFinGer fingerprint $\mathbf{f}_i$ based on a contactless ridge line pattern $\mathbf{r}_{\mathit{cl}} = d(\mathbf{r}_{\mathit{cb}})$ can be formally described as,
\begin{equation*}
\mathbf{f}_i = \mathbf{r}_{\mathit{cl}} + \mathbf{c}_s + \mathbf{c}_{e_i}\mbox{.}
\end{equation*}\vspace{-0.3cm}

	This workflow ensures that SynCoLFinGer is able to robustly create a set of mated samples. Further, the generation of a fingerprint image is highly controllable and easy to extend. 
	
	\subsection{Contactless Capturing} \label{sec:capturing}
\begin{figure}[t]	
		\centering
		\subfigure[Contactless ridge pattern]{\includegraphics[height=4cm]{figs/1_warped_crop}}\hfil
		\subfigure[Rolling rotation]{\includegraphics[height=4cm]{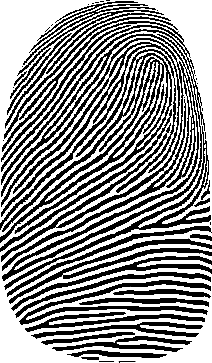}}\hfil
		\subfigure[Ridge line thinning and blur]{\includegraphics[height=4cm]{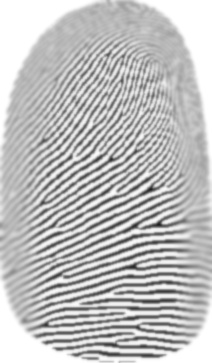}}
		\caption{Processing of the deformed ridge pattern to simulate effects related to the capturing process.}
        \label{fig:contactlessfingerprint}\vspace{-0.3cm}
	\end{figure}

	The simulation of a contactless captured fingerprint includes three processing steps (see Figs. \ref{fig:deformationfigure}, \ref{fig:contactlessfingerprint}): 
	\begin{itemize}
		\item Fingerprint deformation
		\item Rotation
		\item Ridge line thinning and blurring
	\end{itemize}
	
	The fingerprint deformation transforms the SFinGe ridge pattern into a contactless fingerprint ridge pattern. 
	Our deformation method $d$ implements a warp transformation which distorts every pixel $P(x,y)$ along a deformation vector $\Vec{v}$ which starts at $P_s$ (and ends at $P_f$). The intensity of the warping is defined by $||\Vec{v}||$. The warping is implemented by shifting the pixel pixel $P(x,y)$ to its transformed pixel $P'(x,y) = P(x,y) * (||\Vec{v}|| * \log(||\Vec{p}||))$ where $||\Vec{p}||$ is the magnitude between the  $P(x,y)$ and $P_s$.

    The distortion vectors are equally arranged around the contact-based ridge pattern and are facing towards the fingerprint template. The size of the contact-based ridge pattern defines the amount of deformation vectors, the length of the vectors and the distance between the ridge pattern and the vectors. 
	The length of the distortion vectors in the upper area of the fingerprint is higher compared to the left and right border of the finger to represent the natural shape of a human finger.
	Fig.~\ref{fig:deformationfigure} illustrates this process. 
	
	Slight rotations along the rolling axis occur because of the level of freedom during the finger presentation. SynCoLFinGer simulates these rotations by re-using the deformation algorithm. Here, the deformation vectors are facing horizontally in the opposite direction of the rolling. This leads to a stretching of one side and a compression on the other side of the fingerprint template which simulates a rolling of the finger (see  Fig. \ref{fig:contactlessfingerprint} b). Our method simulates a rotation of up to approx. seven degrees of rolling. 
	
	Subsequently, the ridge lines are thinned by an erosion algorithm to achieve a realistic ridge line appearance in the contactless sample. Further, the image is blurred. A Gaussian blur is applied to the whole sample to smooth the transitions between ridges and valleys. The depth-of-field of a contactless capturing device is usually only a few millimeters which causes an out-of-focus blur at the border of the finger \cite{raghavendra2013scaling}. This  property is simulated by a blurring of the finger border regions (see Fig. \ref{fig:contactlessfingerprint} c). 

	\subsection{Subject-related Characteristics}\label{sec:subject} 
\begin{figure}[t]	
		\centering
		\subfigure[Regions of low contrast]{\includegraphics[height=4cm]{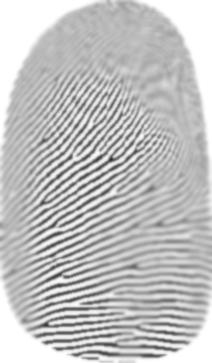}}\hfil
		\subfigure[Finger shape and skin tone]{\includegraphics[height=4cm]{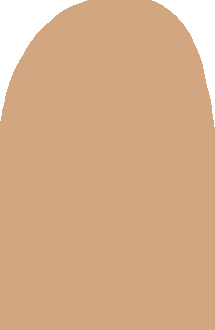}}\hfil
		\subfigure[Skin tone variation]{\includegraphics[height=4cm]{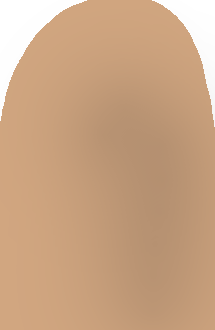}}\hfil
		\caption{Simulation of subject-related characteristics.}
		\label{fig:subject_characteristic}\vspace{-0.3cm}
	\end{figure}

	The generation of realistic contactless fingerprint images requires the simulation of certain subject-specific characteristics. These characteristics refer to (see  Fig. \ref{fig:subject_characteristic}):
	\begin{itemize}
		\item Regions of low contrast
		\item Skin color 
		\item Skin tone variation
	\end{itemize}

	Most contactless fingerprint images show regions of low contrast. These are mainly caused by worn out fingers or dermatological issues. To simulate such properties, local regions of the fingerprint image are blurred (Fig. \ref{fig:subject_characteristic} a). 
	
	To acquire a set of fingertip skin colors we manually analysed a subset of the ISPFDv2 database \cite{MALHOTRA2017119}. More precisely, we randomly selected 25 subjects. For each subject, ten data points within the fingerprint region are averaged to a skin color representation (\cf Fig. \ref{fig:subject_characteristic} b).  
	
	The color of a fingertip is not perfectly homogeneous throughout the whole surface. Skin impurities or damages cause variations of the skin tone. This is simulated by a local brightness and color variations. Here, a simple filter slightly varies the brightness and color of the finger area (Fig. \ref{fig:subject_characteristic} c). 
	
    \subsection{Environmental Influences}\label{sec:environment}
\begin{figure}[t]	
		\centering
		\subfigure[Shadow]{\includegraphics[height=4cm]{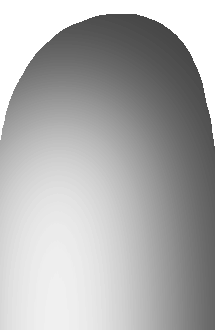}}\hfil
		\subfigure[Inversion of ridge lines]{\includegraphics[height=4cm]{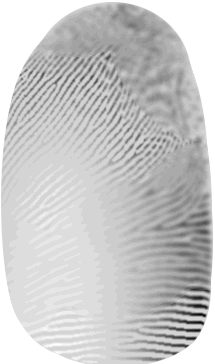}}\\
		\subfigure[Camera noise]{\includegraphics[height=4cm]{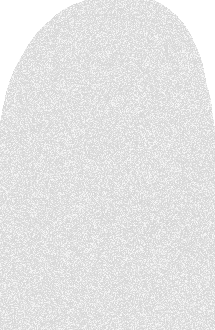}}\hfil
		\subfigure[Dirt]{\includegraphics[height=4cm]{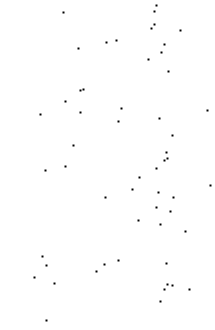}}\hfil
		\subfigure[Finger image]{\includegraphics[height=4cm]{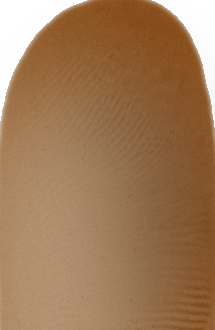}}
		\caption{Simulation of environmental influences like shadow and camera noise.}\vspace{-0.3cm}
		\label{fig:environmentalinfluences}
	\end{figure}

    Numerous environmental influences need to be considered, such as variations in the capturing scenario, technical preconditions of the capturing device, or particles on the finger surface.  
    SynCoLFinGer implements some of the most relevant environmental challenges (\cf Fig. \ref{fig:environmentalinfluences}): 
    \begin{itemize}
        \item Brightness and color variation\vspace{-0.0cm}
		\item Shadow\vspace{-0.0cm}
		\item Dirt\vspace{-0.0cm}
		\item Illumination\vspace{-0.0cm}
		\item Camera noise
	\end{itemize}	
	
	The skin color impression in contactless capturing scenarios also depends on the environment. Here, the illumination of the surrounding area and the background of the image have an influence on the skin tone appearance. This is simulated through a general variation of the skin tone. 
		
	Another aspect which is closely related to illumination is the shadow on the finger area. SynCoLFinGer simulates the shadow as darker regions at the border of a finger image. Here, a shadow mask implements the illumination of the fingertip with a single light source (Fig. \ref{fig:environmentalinfluences} (a)).

	\begin{figure}[t]	
		\centering
		\subfigure[High quality]{\includegraphics[height=0.45\linewidth]{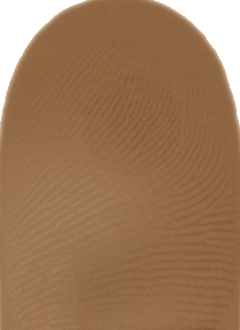}}\hfil
		\subfigure[Medium quality]{\includegraphics[height=0.45\linewidth]{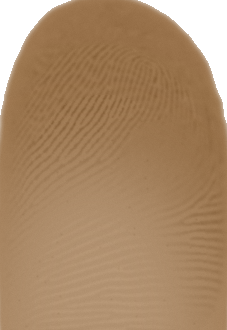}}\hfil
		\subfigure[Low quality]{\includegraphics[height=0.45\linewidth]{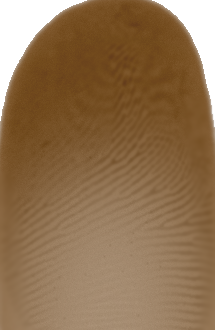}}
		\caption{Three synthetically generated samples of a single finger with different quality presets: (a) high, (b) medium (b) and (c) low quality.}\vspace{-0.3cm}
        \label{fig:intersamplevariance}
	\end{figure}

\begin{figure}[t]	
	\centering
		\subfigure[Real]{
		\resizebox{\linewidth}{!}{
		\includegraphics[height=0.12\linewidth]{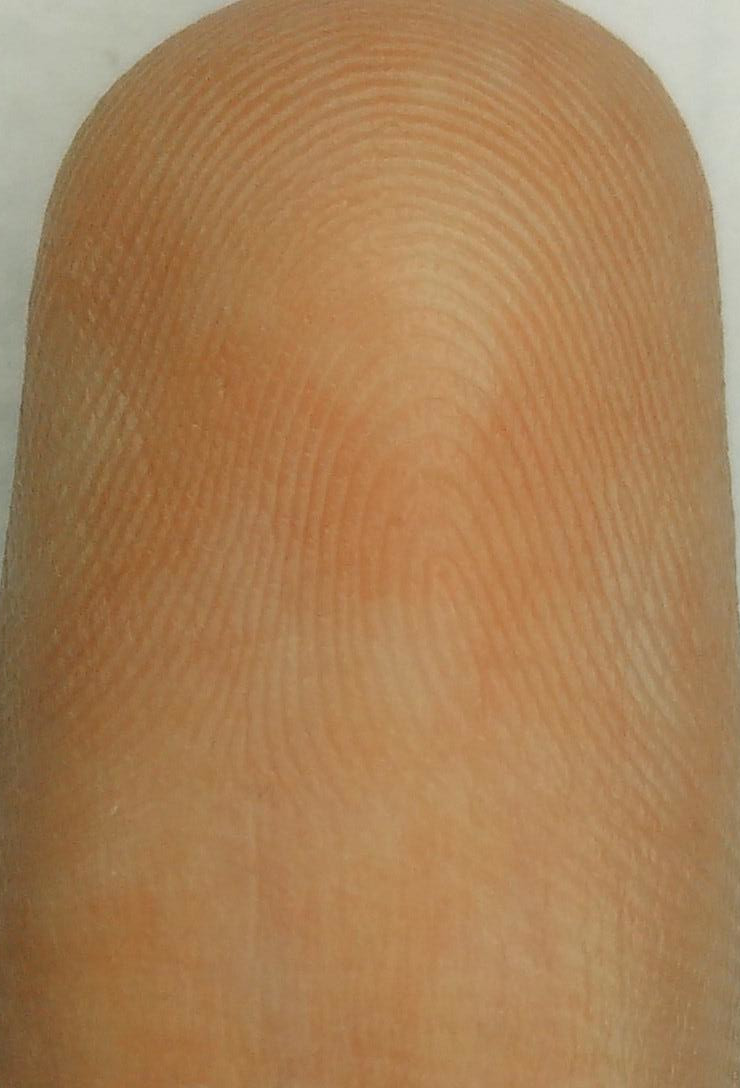}
		\includegraphics[height=0.12\linewidth]{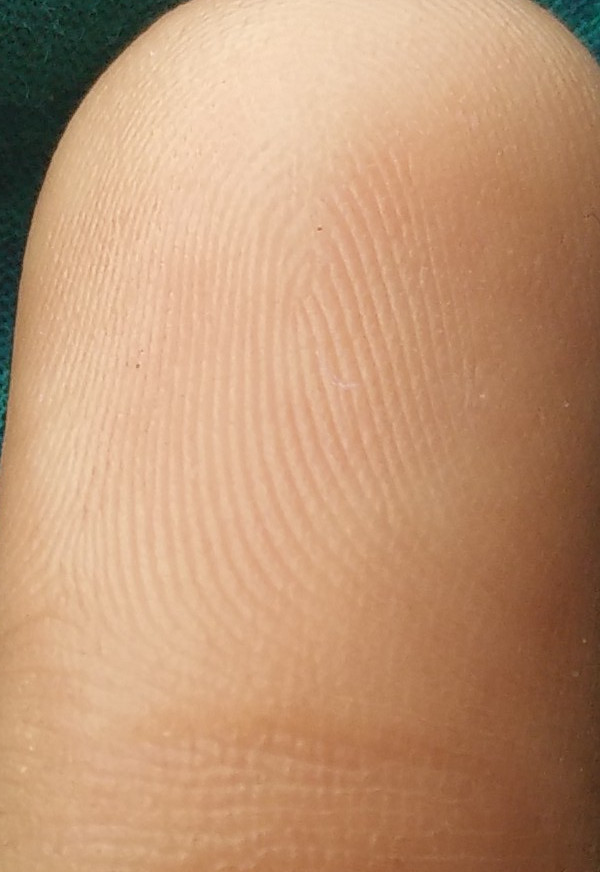}
		\includegraphics[height=0.12\linewidth]{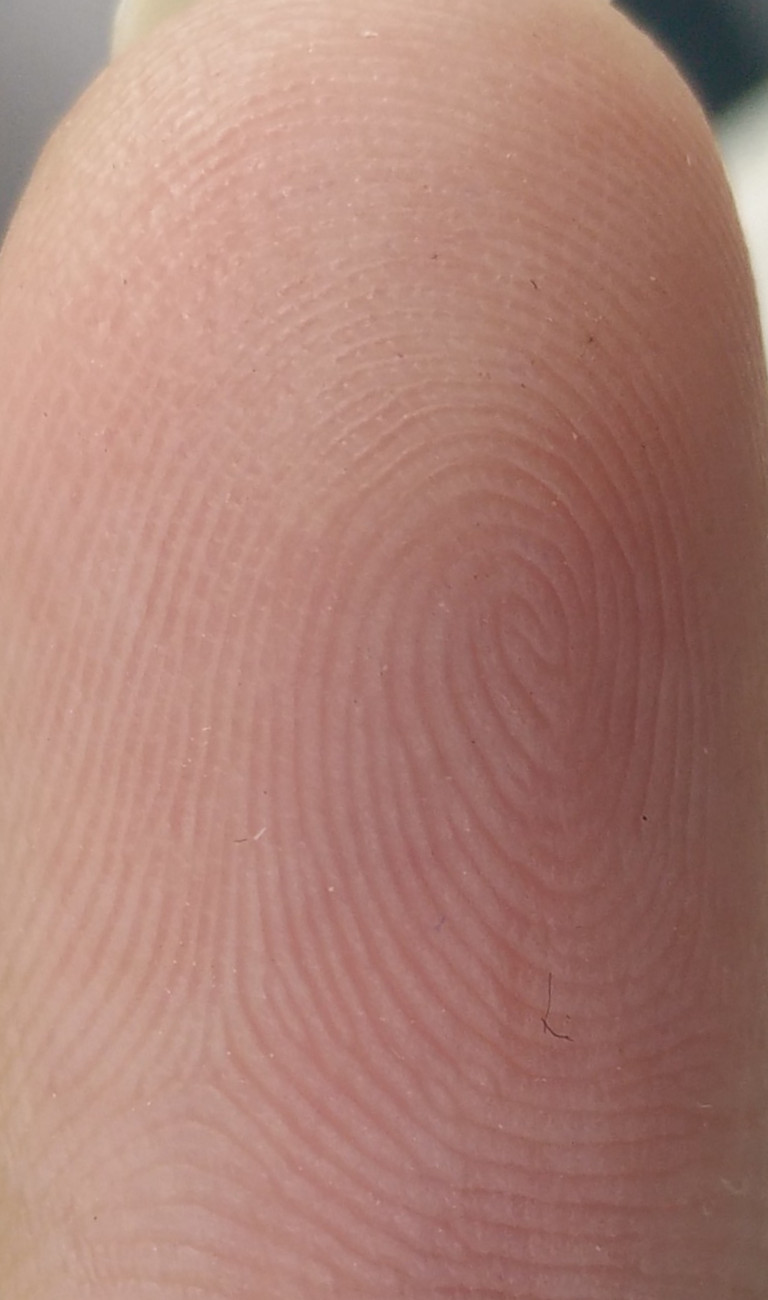}
		}
	}\vspace{-0.2cm}
		\subfigure[Synthetic -- high quality]{
		\resizebox{\linewidth}{!}{
		\includegraphics[height=0.12\linewidth]{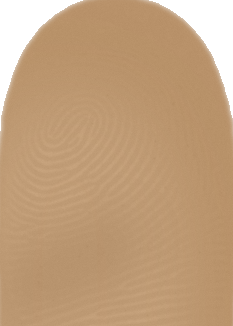}
		\includegraphics[height=0.12\linewidth]{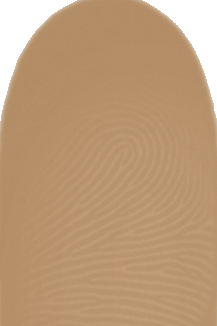}
		\includegraphics[height=0.12\linewidth]{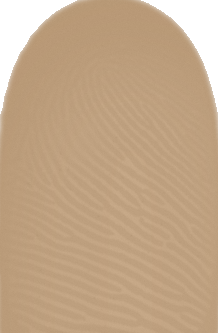}
		}
	}\vspace{-0.2cm}
	\subfigure[Synthetic -- medium quality]{
	\resizebox{\linewidth}{!}{
		\includegraphics[height=0.12\linewidth]{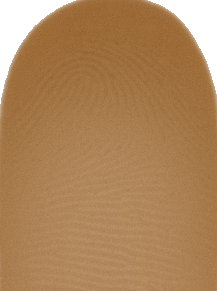}
		\includegraphics[height=0.12\linewidth]{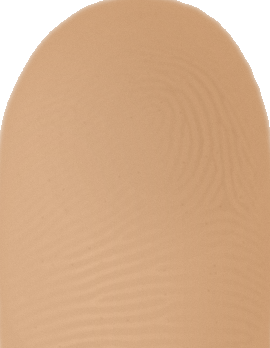}
		\includegraphics[height=0.12\linewidth]{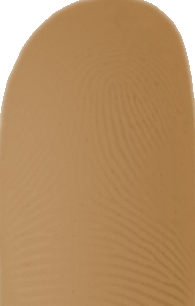}

	}
	}\vspace{-0.2cm}
	\subfigure[Synthetic -- low quality]{
	\resizebox{0.95\linewidth}{!}{
		\includegraphics[height=0.12\linewidth]{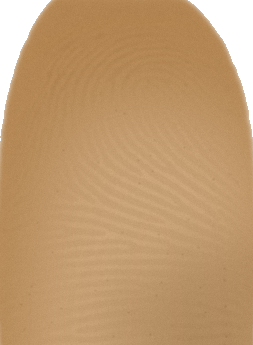}
		\includegraphics[height=0.12\linewidth]{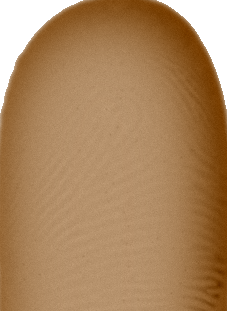}
		\includegraphics[height=0.12\linewidth]{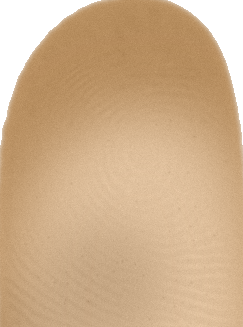}}
	}\vspace{-0.2cm}
	\caption{Examples of real and synthetic contactless fingerprint images generated by SynCoLFinGer: (a) real images, synthetic images of (b)  high, (c) medium and (d) low quality.}
	\label{fig:finger_examples}\vspace{-0.3cm}
\end{figure}

    Ridge line inversion refers to the switching of parts of the ridge line structure from bright to dark because of illumination and reflection properties of the skin. In brighter regions the ridge lines appear brighter than the fingertip whereas in dark regions the ridge lines appear darker (Fig. \ref{fig:environmentalinfluences} (b)). The fingerprint sample is adjusted accordingly to simulate this property. For this purpose, the shadow mask is used to have an indication of bright and dark regions in the finger area. 
  
	Camera noise is directly related to illumination and the capturing device. The adding of camera noise simulates a capturing device which operates under challenging light conditions (Fig. \ref{fig:environmentalinfluences} (c)). During the simulation, the illumination influences the amount of noise generated, whereas darker preconditions lead to more noise.
    
    Dirt refers to temporary particles on the finger surface which may influence the recognition performance. SynCoLFinGer implements the presence of dirt by particles which are randomly distributed over the finger region (Fig. \ref{fig:environmentalinfluences} (d)).
	Finally, slight variations in terms of rotation and scaling are applied to every sample in order to simulated inaccuracies during the segmentation.

	The different masks have different properties in terms of persistence. Some properties like the skin color or the fingerprint characteristic itself are highly persistent, whereas others like illumination may change more drastically. Some properties like regions of low contrast caused by worn-out fingers are semi-persistent and change after days or weeks. 
	For this reason semi-persistent masks are also equipped with an alteration factor which may represent the time between two capturing sessions.

	\section{Evaluation}\label{sec:experiments}
	In this section we evaluate the suitability of our proposed method in a biometric recognition scenario. First, we visually inspect the generated database and discuss its properties. Second, we evaluate the amount of detected minutiae, the sample quality and the biometric recognition performance. Here, we used the workflow proposed in one of our previous works \cite{priesnitz2021mobile} as described below. 
	
	\begin{figure}[t]	
	\centering
	\subfigure[ISPFDv2]{\includegraphics[height=4.8cm]{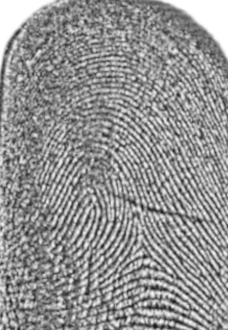}}\hfill
	\subfigure[SynCoLFinGer]{\includegraphics[height=4.8cm]{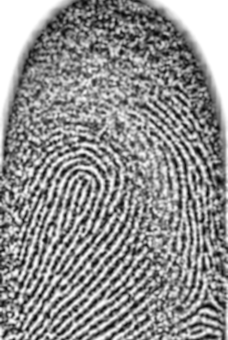}}
	\caption{Examples of a real and synthetic contactless fingerprint sample pre-processed with our method: (a) SynCoLFinGer, (b) contactless sample from the ISPFDv2 database}
	\label{fig:fingerprint_prep}\vspace{-0.3cm}
\end{figure}

\begin{figure*}[t]	
	\centering
	\subfigure[High quality]{\includegraphics[height=3.5cm]{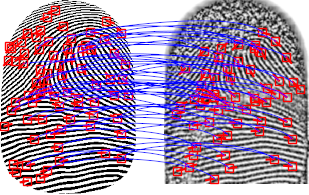}}\hfill
	\subfigure[Medium quality]{\includegraphics[height=3.5cm]{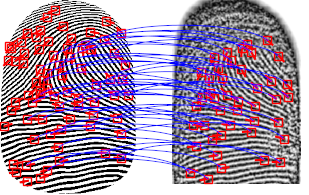}}\hfill
	\subfigure[Low quality]{\includegraphics[height=3.5cm]{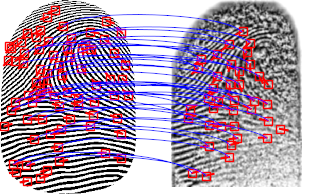}}
	\caption{Minutiae correspondence between the contactless ridge pattern (\cf Fig.~3b) (left) and the pre-processed generated contactless fingerprint (right) of (a) high, (b) medium and (c) low quality.}\label{fig:minutiae_correspondence}\vspace{-0.3cm}
\end{figure*}
	
	\subsection{Synthetic Database Generation}
	To evaluate our algorithm, we generate a database of synthetic contactless fingerprint samples. To show the achievable variation three subsets are created: 
	\begin{itemize}
		\item \emph{High}: samples which are captured under good conditions, \ie samples of high  quality.\vspace{-0.0cm}
		\item \emph{Medium}: samples which are captured under good conditions but where the fingertip is slightly rotated or the fingerprint has a low contrast. \vspace{-0.0cm}
		\item \emph{Low}: challenging samples with huge regions of low contrast, strong rotation exhibiting huge variation in terms of illumination, \ie samples of low  quality. 
	\end{itemize}
	In addition, changeable properties described in the previous section are varied randomly to obtain several mated contactless fingerprint samples from a single ridge pattern. Fig.~\ref{fig:intersamplevariance} illustrates an example of the above described quality variations for mated samples. The generated database consists of $1,000$ different fingerprints (SFinGe samples). For each category and fingerprint six samples are generated which results in a total amount of $18,000$ samples.

    \subsection{Evaluation Pipeline}
    Our generated samples are analysed in terms of biometric sample quality and recognition performance. To this end, we apply a pre-processing to the generated images, evaluate  biometric sample quality using the NFIQ 2.0 algorithm and measure the biometric utility using a state-of-the-art contactless fingerprint recognition system.  
    To enable a processing of contactless data with algorithms designed for the contact-based domain, a pre-processing has to be applied which transforms a contactless fingerprint image to a contact-based equivalent fingerprint image \cite{8296943}. 
    The pre-processing consists of a grayscale conversion, image width normalization and a Contrast Limited Adaptive Histogram Equalization (CLAHE). For a more detailed description of the applied method the reader is referred to \cite{priesnitz2021mobile}.    Fig.~\ref{fig:fingerprint_prep} depicts a real and a synthetic fingerprint image pre-precessed by our proposed method.  With this pre-processing it was shown that NFIQ 2.0 is able to accurately predict the quality of a contactless sample \cite{Priesnitz-Fingerprint-Quality-BIOSIG-2020}.

It should be noted that the fingerprint images generated by SynCoLFinGer are segmented, cropped to the ROI and rotated in an upright position. For this reason, these processing steps can be skipped. On a database captured under unconstrained environmental influences these steps also have to be applied. 

To measure the biometric recognition performance in terms of the standardised metrics False Match Rate (FMR), False Non-Match Rate (FNMR) \cite{ISO-IEC-19795-1-Framework-210216}, and Equal Error Rate (EER) we combine the FingerNet method of Tang \etal \cite{Tang-FingerNet-2017} for feature extraction with the comparison algorithm of Va\v{z}an \etal \cite{Vazan-SourceAFIS-2019}. The feature extractor is based on deep-learning and outputs minutiae features (positions and orientations). Then a comparison score is computed based on a nearest neighbour algorithm.  

	\subsection{Experimental Results}
	Requirements for the assessment of synthetic fingerprint generation have recently been defined by Makrushin~\etal~\cite{Makrushin21}. Stipulated properties include data anonymity, sufficiently high image resolution, diversity and uniqueness. These requirements are fulfilled through the use of the SFinGe algorithm. The requirement of controllable generation is met by the conceptual design of SynCoLFinGer described in the previous section.
	
	Realistic appearance represents another key requirement for synthetic fingerprint images \cite{Makrushin21}. A visual inspection of the generated database reveals that the SynCoLFinGer samples show a good level of realism. Examples of synthetically generated images are depicted in Fig.~\ref{fig:finger_examples}. Important aspects like a realistic impression of the ridge pattern, regions of low contrast, skin tone, and environmental influences are represented in the generated samples. Moreover, the biometric variation between mated samples  realistically simulates the capturing sessions under different conditions, \cf Fig.~\ref{fig:intersamplevariance}. Because of the controllable design of the algorithm, generated samples show realistic ridge patterns, in contrast to some recently proposed GAN-based approaches for generating contact-based fingerprints.
	Nevertheless, it should be noted that the generated images can still be distinguished from real ones (\cf Fig.~\ref{fig:finger_examples}). Note that this is also caused by the missing background of the generated fingerprint images.

	We also investigate the appearance of the synthetic samples after applying our pre-processing pipeline. Fig.~\ref{fig:fingerprint_prep} compares a fingerprint from the ISPFDv2 database with a sample generated by SynCoLFinGer. It should be noted that the same automatic pre-processing pipeline was applied which was originally designed for real contactless fingerprints. The applied pre-processing pipeline robustly emphasizes the ridgeline pattern of real and generated contactless fingerprint images.

	An important aspect for the level of realism of the generated images is the number of minutiae extracted by the feature extractor. To analyse this, we use the FingerNet algorithm \cite{Tang-FingerNet-2017} for feature extraction and count the average amount of detected minutiae per fingerprint image. From Table~\ref{tab:results} we can see that the amount of minutiae slightly drops from high over medium to the low quality subset. Also, our generated database aligns with the real databases in terms of minutiae count. We  can also observe from the table that a large amount of detected minutiae does not necessarily lead to an improved recognition performance as shown on the HDA database. This can result from the detection of false minutiae. Also, scaling and rotation of fingerprints might be necessary to lower the EER in such cases.

	Moreover, we performed a minutiae correspondence analysis. Fig.~\ref{fig:minutiae_correspondence} shows a contactless ridge pattern and its corresponding processed sample from the low quality sub-database. The red rectangles represent minutiae detected by the FingerNet method. The blue connections indicate corresponding minutiae. It is observable that many minutiae from the contactless ridge pattern are also represented in pre-processed sample. Especially in the center region, many minutiae from the ridge pattern are also extracted from the contactless sample. Nevertheless, minutiae at the border region are often not detected in the pre-processed sample due to the simulated out-of-focus blur. It should be noted that in Fig.~\ref{fig:minutiae_correspondence} the relative position of minutiae is shifted in the pre-processed samples of the medium quality image (Fig.~\ref{fig:minutiae_correspondence} (b)) and the low quality image (Fig.~\ref{fig:minutiae_correspondence} (c)). This  is caused by a rotation of the fingertip during the generation. 
	These observations are highly comparable to real databases, \eg the ISPFDv2 database.

	\begin{table*}[!t]
		\centering
		\caption{Average NFIQ 2.0 scores and biometric performance obtained from the synthetically generated database compared to real contactless fingerprint databases.}\label{tab:results}
		\resizebox{0.7\linewidth}{!}{
		\begin{small}
			\begin{tabular}{|c|c|c|c|c|}
				\hline
				\textbf{DB} & \textbf{Subset} & \textbf{Minutiae count} & \textbf{NFIQ 2.0 score} & \textbf{EER (\%)} \\\hline
				\multirow{3}{*}{SynCoLFinGer} & high  & 42.82 ($\pm$ 19.93) & 40.68 ($\pm$7.93)  & 1.13
				 \\\cline{2-5}
				& medium & 37.90 ($\pm$ 10,99) & 34.92 ($\pm$10.04)  & 3.07
				 \\\cline{2-5}
				& low & 36.06 ($\pm$ 11.37) &  27.80 ($\pm$12.66) & 6.44
				 \\\Xhline{2\arrayrulewidth}
				\multirow{2}{*}{PolyU \cite{8244291}} & session 1 & 37.09 ($\pm$ 10.74) & 47.71 ($\pm$10.86)  & 3.91 \\\cline{2-5}
				& session 2 & 36.17 ($\pm$ 11.84) & 47.08 ($\pm$13.21)  & 3.17 \\\Xhline{2\arrayrulewidth}
				\multirow{2}{*}{ISPFDv1 \cite{MALHOTRA2017119}} & white & 48.01 ($\pm$ 17.91) & 17.73 ($\pm$8.40) & 30.48 \\\cline{2-5}
				& natural & 49.84 ($\pm$ 20,53) & 12.16 ($\pm$8.52) & 31.38 \\\cline{2-5} \Xhline{2\arrayrulewidth} 
				\multirow{2}{*}{HDA \cite{priesnitz2021mobile}} & constrained & 51.21 ($\pm$ 29.56) & 44.80 ($\pm$12.36) & 11.16\\\cline{2-5}
				& unconstrained & 69.26 ($\pm$ 24.98) & 36.13 ($\pm$14.05) &  29.75 \\\hline
			\end{tabular}
		\end{small}
		}
	\end{table*}	
	
\begin{figure}
		\centering
		\includegraphics[width=0.999\linewidth]{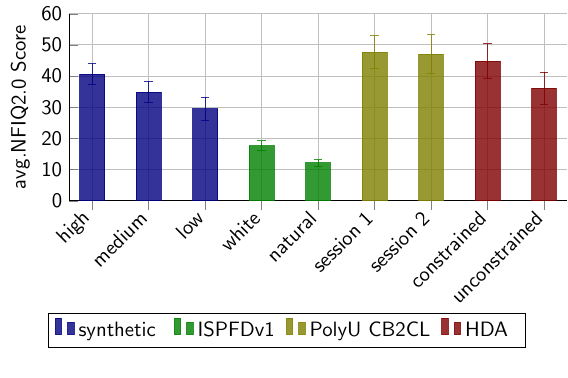}
		\vspace{-0.7cm}
		\caption{Overview of NFIQ 2.0 scores obtained from different contactless databases: synthetic fingerprint image generated by SynCoLFinGer (blue), the contactless sub-databases of the ISPFDv1 database (green), the contactless sub-databases of the PolyU database (yellow) and a database acquired with our own capturing device (red).}
		\label{fig:nfiqoverview}\vspace{-0.3cm}
	\end{figure}

	In the next step, we investigate the most important requirement on whether the generated fingerprint images reflect the basic characteristics of real contactless fingerprint data. 
	Average obtained NFIQ 2.0 scores for synthetically generated fingerprint images at different quality level are depicted in  Fig.~\ref{fig:nfiqoverview} and listed in Table~\ref{tab:results} along with a comparison to NFIQ 2.0 scores achieved on popular real datasets.  It can be observed that the sample quality of the SynCoLFinGer images is within the range of real contactless fingerprint databases or slightly better.  As expected, the three categories are showing different biometric accuracy. The high-quality samples have an average NFIQ 2.0 score above 40. For medium-quality samples, the average NFIQ 2.0 score is dropping by approx. 5 points. The lowest NFIQ 2.0 score of approximately 28 is obtained on the low-quality samples. The standard deviation of NFIQ 2.0 scores is in a similar range compared to the real databases. This shows that SynCoLFinGer is able to produce a similar amount of variance in terms of sample quality. 
	We see that the high-quality database has the lowest standard deviation, whereas the low-quality category has the highest. This represents the parameter settings in our configuration files and can be adapted easily. Real contactless databases contain samples with different ridge line frequency, especially if they are captured in unconstrained setups (\eg "ISPFDv1 natural" and "HDA unconstrained"). Databases captured in more constrained scenarios show much higher NFIQ 2.0 scores (\eg "PolyU" or "HDA constrained"). Especially, the distance between the capturing device and the finger needs to be kept constant to approximate a fingerprint size which is equivalent to 500 dpi contact-based fingerprints (which is expected by the NFIQ 2.0 algorithm). The SFinGe templates on which our method is based simulate a 500 dpi capturing device which is also favored by NFIQ 2.0. Even if the samples are highly deformed, the size of a fingerprint and the ridge line frequency in the central area of the fingerprint contribute to a high NFIQ 2.0 score. It is noteworthy that NFIQ 2.0 was found to be suboptimal for biometric sample quality assessment on contactless fingerprint data and, hence, its application requires specific pre-processing \cite{Priesnitz-Fingerprint-Quality-BIOSIG-2020}. Alternatively, quality assessment algorithms specifically designed for contactless fingerprint data could be employed, \eg \cite{5596694,6595867}.

    \begin{figure}
    	\centering
    	\includegraphics[width=0.999\linewidth]{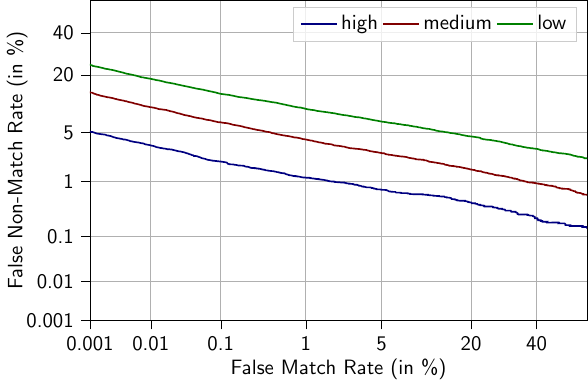}
    	\caption{Detection Error Trade-off (DET) curves for synthetically generated contactless fingerprint images at different sample quality levels.}
    	\label{fig:examplealgorithm}\vspace{-0.3cm}
    \end{figure}
    
	The evaluation in terms of biometric recognition performance shows that SynCoLFinGer samples are working well in an open-source contactless fingerprint recognition workflow (see Table~\ref{tab:results}). As expected, the three categories are showing different biometric accuracy. The high, medium, and low-quality samples reveal EERs of 1.13\%, 3.07\%, and 6.44\%, respectively. This desired behaviour is directly caused by the parameter settings in the configuration presets and is observable across different decision threshold configurations as shown in the DET plots of Fig.~\ref{fig:examplealgorithm}. It is observable that especially databases acquired in unconstrained conditions (\eg "ISPFDv1 natural" and "HDA unconstrained") have much higher EERs compared to the SynCoLFinGer samples. A possible explanation for this is a slight variance of the segmented fingertip, which results in a shifted minutiae map. Because the SynCoLFinGer samples are not suffering from this variance the recognition accuracy is expected to be much higher. 
	

	Overall, the equal error rates and NFIQ 2.0 scores show that SynCoLFinGer is able to generate synthetic contactless fingerprint images which closely reflect the  characteristics of real contactless fingerprint data in terms of sample quality and biometric utility.

	\section{Conclusion}\label{sec:conclusion}
	This work introduced SynCoLFinGer, a synthetic contactless fingerprint generator. To the best of our knowledge, SynCoLFinGer represents the first approach towards the generation of synthetic contactless fingerprint images. In the proposed method, properties of contactless fingerprints with respect to the capturing process, subject-related characteristics, and environmental influences were simulated.
	It was shown that the proposed method is able to generate synthetic mated samples at various quality levels. In experiments, it was demonstrated that the synthetically generated contactless fingerprints reflect the basic characteristics of real contactless fingerprint data in terms of biometric sample quality and biometric utility.  
	
		Our proposed approach is expected to open new possibilities for further contributions in which different avenues for future research could be considered, \textit{e.g.} an improvement of realism of the generated contactless fingerprint images including more environmental variations such as drastic changes in image resolution or the generation of synthetic contact-based and contactless samples for interoperability studies.

\section*{Acknowledgements}
\label{sec:acknowledgements}
This research work has been funded by the German Federal Ministry of Education and Research and the Hessian Ministry of Higher Education, Research, Science and the Arts within their joint support of the National Research Center for Applied Cybersecurity ATHENE.

\bibliographystyle{IEEEtran}
\bibliography{references}

\begin{thebibliography}{10}
\providecommand{\url}[1]{#1}
\csname url@samestyle\endcsname
\providecommand{\newblock}{\relax}
\providecommand{\bibinfo}[2]{#2}
\providecommand{\BIBentrySTDinterwordspacing}{\spaceskip=0pt\relax}
\providecommand{\BIBentryALTinterwordstretchfactor}{4}
\providecommand{\BIBentryALTinterwordspacing}{\spaceskip=\fontdimen2\font plus
\BIBentryALTinterwordstretchfactor\fontdimen3\font minus
  \fontdimen4\font\relax}
\providecommand{\BIBforeignlanguage}[2]{{%
\expandafter\ifx\csname l@#1\endcsname\relax
\typeout{** WARNING: IEEEtran.bst: No hyphenation pattern has been}%
\typeout{** loaded for the language `#1'. Using the pattern for}%
\typeout{** the default language instead.}%
\else
\language=\csname l@#1\endcsname
\fi
#2}}
\providecommand{\BIBdecl}{\relax}
\BIBdecl

\bibitem{Buettner2009}
D.~J. Buettner, ``Biometric sample synthesis,'' in \emph{Encyclopedia of
  Biometrics}, S.~Z. Li and A.~Jain, Eds.\hskip 1em plus 0.5em minus
  0.4em\relax Springer, 2009, pp. 116--122.

\bibitem{labati2012virtual}
R.~D. Labati, A.~Genovese, V.~Piuri, and F.~Scotti, ``Virtual environment for
  3-d synthetic fingerprints,'' in \emph{2012 IEEE International Conference on
  Virtual Environments Human-Computer Interfaces and Measurement Systems
  (VECIMS) Proceedings}.\hskip 1em plus 0.5em minus 0.4em\relax IEEE, 2012, pp.
  48--53.

\bibitem{long20153d}
S.~Long, S.~Li, Q.~Zhao, and W.~Song, ``3d fingerprint modelling and
  synthesis,'' \emph{Electronics Letters}, vol.~51, no.~18, pp. 1418--1420,
  2015.

\bibitem{Cappelli2009}
R.~Cappelli, ``Sfinge,'' in \emph{Encyclopedia of Biometrics}, S.~Z. Li and
  A.~Jain, Eds.\hskip 1em plus 0.5em minus 0.4em\relax Springer, 2009, pp.
  1169--1176.

\bibitem{Hillerstroem14}
F.~Hillerström, A.~Kumar, and R.~Veldhuis, ``Generating and analyzing
  synthetic finger vein images,'' in \emph{Int'l Conf. of the Biometrics
  Special Interest Group (BIOSIG'14)}, 2014, pp. 1--9.

\bibitem{Goodfellow14}
I.~Goodfellow, J.~P.-A.~M. Mirza, B.~Xu, D.~Warde-Farley, S.~Ozair,
  A.~Courville, and Y.~Bengio, ``Generative adversarial nets,'' in
  \emph{Advances in Neural Information Processing Systems ({NIPS})}, 2014.

\bibitem{Karras19}
T.~Karras, S.~Laine, and T.~Aila, ``A style-based generator architecture for
  generative adversarial networks,'' in \emph{Conf. on Computer Vision and
  Pattern Recognition (CVPR'19)}, 2019, pp. 4396--4405.

\bibitem{Yadav19}
S.~Yadav, C.~Chen, and A.~Ross, ``Synthesizing iris images using rasgan with
  application in presentation attack detection,'' in \emph{Conf. on Computer
  Vision and Pattern Recognition Workshops (CVPRW'19)}, 2019, pp. 2422--2430.

\bibitem{engelsma2022printsgan}
J.~J. Engelsma, S.~A. Grosz, and A.~K. Jain, ``Printsgan: Synthetic fingerprint
  generator,'' \emph{arXiv preprint arXiv:2201.03674}, 2022.

\bibitem{seidlitz2021generation}
S.~Seidlitz, K.~J{\"u}rgens, A.~Makrushin, C.~Kraetzer, and J.~Dittmann,
  ``Generation of privacy-friendly datasets of latent fingerprint images using
  generative adversarial networks.'' in \emph{VISIGRAPP (4: VISAPP)}, 2021, pp.
  345--352.

\bibitem{Maltoni-HandbookOfFingerprintRecognition-2009}
D.~Maltoni, D.~Maio, A.~Jain, and S.~Prabhakar, \emph{Handbook of Fingerprint
  Recognition}, 1st~ed.\hskip 1em plus 0.5em minus 0.4em\relax Springer-Verlag,
  2009.

\bibitem{Cappelli20}
R.~Cappelli, A.~Erol, D.~Maio, and D.~Maltoni, ``Synthetic fingerprint-image
  generation,'' in \emph{Int'l Conf. on Pattern Recognition. (ICPR'00)}, 2000,
  pp. 471--474 vol.3.

\bibitem{Zhao12}
Q.~Zhao, A.~K. Jain, N.~G. Paulter, and M.~Taylor, ``Fingerprint image
  synthesis based on statistical feature models,'' in \emph{Int'l Conf. on
  Biometrics: Theory, Applications and Systems (BTAS'12)}, 2012, pp. 23--30.

\bibitem{riazi2020synfi}
M.~S. Riazi, S.~M. Chavoshian, and F.~Koushanfar, ``Synfi: Automatic synthetic
  fingerprint generation,'' \emph{arXiv preprint arXiv:2002.08900}, 2020.

\bibitem{Fahim20}
M.~A.-N.~I. Fahim and H.~Y. Jung, ``A lightweight gan network for large scale
  fingerprint generation,'' \emph{IEEE Access}, vol.~8, pp. 92\,918--92\,928,
  2020.

\bibitem{drozdowski2020demographic}
P.~Drozdowski, C.~Rathgeb, A.~Dantcheva, N.~Damer, and C.~Busch, ``Demographic
  bias in biometrics: A survey on an emerging challenge,'' \emph{IEEE
  Transactions on Technology and Society}, vol.~1, no.~2, pp. 89--103, 2020.

\bibitem{labati2014touchless}
R.~D. Labati, A.~Genovese, V.~Piuri, and F.~Scotti, ``Touchless fingerprint
  biometrics: A survey on 2d and 3d technologies,'' \emph{Journal of Internet
  Technology}, vol.~15, no.~3, p. 328, 2014.

\bibitem{priesnitz2021overview}
J.~Priesnitz, C.~Rathgeb, N.~Buchmann, C.~Busch, and M.~Margraf, ``An overview
  of touchless 2d fingerprint recognition,'' \emph{EURASIP Journal on Image and
  Video Processing}, vol. 2021, no.~1, pp. 1--28, 2021.

\bibitem{priesnitz2021deep}
J.~Priesnitz, C.~Rathgeb, N.~Buchmann, and C.~Busch, ``Deep learning-based
  semantic segmentation for touchless fingerprint recognition,'' in
  \emph{International Conference on Pattern Recognition}.\hskip 1em plus 0.5em
  minus 0.4em\relax Springer, 2021, pp. 154--168.

\bibitem{raghavendra2013scaling}
R.~Raghavendra, C.~Busch, and B.~Yang, ``Scaling-robust fingerprint
  verification with smartphone camera in real-life scenarios,'' in \emph{2013
  IEEE Sixth International Conference on Biometrics: Theory, Applications and
  Systems (BTAS)}.\hskip 1em plus 0.5em minus 0.4em\relax IEEE, 2013, pp. 1--8.

\bibitem{MALHOTRA2017119}
A.~Malhotra, A.~Sankaran, A.~Mittal, M.~Vatsa, and R.~Singh, ``Fingerphoto
  authentication using smartphone camera captured under varying environmental
  conditions,'' in \emph{Human Recognition in Unconstrained Environments},
  M.~D. Marsico, M.~Nappi, and H.~Proença, Eds.\hskip 1em plus 0.5em minus
  0.4em\relax Academic Press, 2017, pp. 119 -- 144.

\bibitem{priesnitz2021mobile}
J.~Priesnitz, R.~Huesmann, C.~Rathgeb, N.~Buchmann, and C.~Busch, ``{M}obile
  {C}ontactless {F}ingerprint {R}ecognition: {I}mplementation, {P}erformance
  and {U}sability {A}spects,'' \emph{Sensors}, vol.~22, no.~3, 2022.

\bibitem{8296943}
P.~{Salum}, D.~{Sandoval}, A.~{Zaghetto}, B.~{Macchiavello}, and C.~{Zaghetto},
  ``Touchless-to-touch fingerprint systems compatibility method,'' in
  \emph{2017 IEEE International Conference on Image Processing (ICIP)}, 2017,
  pp. 3550--3554.

\bibitem{Priesnitz-Fingerprint-Quality-BIOSIG-2020}
J.~Priesnitz, C.~Rathgeb, N.~Buchmann, and C.~Busch, ``{T}ouchless
  {F}ingerprint {S}ample {Q}uality: {P}rerequisites for the {A}pplicability of
  {NFIQ}2.0,'' in \emph{Proc. Intl. Conf. of the Biometrics Special Interest
  Group ({BIOSIG})}, 2020.

\bibitem{ISO-IEC-19795-1-Framework-210216}
{ISO/IEC JTC1 SC37 Biometrics}, \emph{{ISO/IEC} 19795-1:2021. Information
  Technology -- Biometric Performance Testing and Reporting -- Part~1:
  Principles and Framework}, International Organization for Standardization,
  June 2021.

\bibitem{Tang-FingerNet-2017}
Y.~Tang, F.~Gao, J.~Feng, and Y.~Liu, ``{FingerNet}: An unified deep network
  for fingerprint minutiae extraction,'' in \emph{International Joint
  Conference on Biometrics ({IJCB})}.\hskip 1em plus 0.5em minus 0.4em\relax
  IEEE, October 2017, pp. 108--116.

\bibitem{Vazan-SourceAFIS-2019}
R.~Va\v{z}an, ``{SourceAFIS} -- opensource fingerprint matcher,''
  \url{https://sourceafis.machinezoo.com/}, 2019, last accessed: \today.

\bibitem{Makrushin21}
A.~Makrushin, C.~Kauba, S.~Kirchgasser, S.~Seidlitz, C.~Kraetzer, A.~Uhl, and
  J.~Dittmann, ``General requirements on synthetic fingerprint images for
  biometric authentication and forensic investigations,'' in \emph{Proceedings
  of the 2021 ACM Workshop on Information Hiding and Multimedia Security}, ser.
  IH\&MMSec '21.\hskip 1em plus 0.5em minus 0.4em\relax ACM, 2021, p. 93–104.

\bibitem{5596694}
R.~D. {Labati}, V.~{Piuri}, and F.~{Scotti}, ``Neural-based quality measurement
  of fingerprint images in contactless biometric systems,'' in \emph{The 2010
  International Joint Conference on Neural Networks (IJCNN)}, 2010, pp. 1--8.

\bibitem{6595867}
G.~{Li}, B.~{Yang}, M.~A. {Olsen}, and C.~{Busch}, ``Quality assessment for
  fingerprints collected by smartphone cameras,'' in \emph{2013 IEEE Conference
  on Computer Vision and Pattern Recognition Workshops}, 2013, pp. 146--153.

\bibitem{8244291}
C.~Lin and A.~Kumar, ``Matching contactless and contact-based conventional
  fingerprint images for biometrics identification,'' \emph{IEEE Transactions
  on Image Processing}, vol.~27, no.~4, pp. 2008--2021, 2018.

\end{thebibliography}

\end{document}